\documentclass[letterpaper, 10 pt, conference]{ieeeconf}  % Comment this line out if you need a4paper

\IEEEoverridecommandlockouts                              % This command is only needed if 
\usepackage{graphicx} % for pdf, bitmapped graphics files
\usepackage{epsfig} % for postscript graphics files
\usepackage{mathptmx} % assumes new font selection scheme installed
\usepackage{times} % assumes new font selection scheme installed
\usepackage{amsmath, bm} % assumes amsmath package installed
\usepackage{amssymb}  % assumes amsmath package installed
\usepackage{hyperref}
\usepackage[capitalize]{cleveref}
\usepackage{array}
\usepackage{multirow}
\usepackage{tikz}
\usepackage{url}
\usepackage[T1]{fontenc}
\usepackage{comment}
\usepackage{tensor}
\usepackage{commands}
\usepackage{todonotes}

\title{\LARGE \bf
Reformulating AI-based Multi-Object Relative State Estimation for Aleatoric Uncertainty-based Outlier Rejection of Partial Measurements
}

\author{Thomas Jantos$^{1}$, Giulio Delama$^{1}$, Stephan Weiss$^{1}$ and Jan Steinbrener$^{1}$% <-this % stops a space
\thanks{$^{1}$The authors are with the Control of Networked Systems Group, University of Klagenfurt, 9020 Klagenfurt am Wörthersee, Austria}%
\thanks{This work was supported by the Christian Doppler Forschungsgesellschaft within the project AIONIC.}%
\thanks{{\textbf{Pre-print version, accepted Jan/2026, DOI follows ASAP ~\copyright IEEE.}}}%
}

\begin{document}

\maketitle
\thispagestyle{empty}
\pagestyle{empty}

%%%%%%%%%%%%%%%%%%%%%%%%%%%%%%%%%%%%%%%%%%%%%%%%%%%%%%%%%%%%%%%%%%%%%%%%%%%%%%%%
\begin{abstract}
Precise localization with respect to a set of objects of interest enables mobile robots to perform various tasks. With the rise of edge devices capable of deploying deep neural networks (DNNs) for real-time inference, it stands to reason to use artificial intelligence (AI) for the extraction of object-specific, semantic information from raw image data, such as the object class and the relative six degrees of freedom (6-DoF) pose. However, fusing such AI-based measurements in an Extended Kalman Filter (EKF) requires quantifying the DNNs' uncertainty and outlier rejection capabilities.

This paper presents the benefits of reformulating the measurement equation in AI-based, object-relative state estimation. By deriving an EKF using the direct object-relative pose measurement, we can decouple the position and rotation measurements, thus limiting the influence of erroneous rotation measurements and allowing partial measurement rejection. Furthermore, we investigate the performance and consistency improvements for state estimators provided by replacing the fixed measurement covariance matrix of the 6-DoF object-relative pose measurements with the predicted aleatoric uncertainty of the DNN.
\end{abstract}

%%%%%%%%%%%%%%%%%%%%%%%%%%%%%%%%%%%%%%%%%%%%%%%%%%%%%%%%%%%%%%%%%%%%%%%%%%%%%%%%
\section{Introduction}

Object-relative state estimation enables mobile robots to localize and navigate with respect to objects of interest, crucial for tasks such as object following and critical infrastructure inspection \cite{jantos2024aivio}. In previous work, Jantos et al. \cite{jantos2023aiobjrelstate} introduced an extended Kalman filter (EKF)-based approach for object-relative state estimation, concurrently estimating the state of the mobile robot and the pose of the objects with respect to the navigation frame. An inertial measurement unit (IMU) is used as the propagation sensor and fused with artificial intelligence (AI)-based six degrees of freedom (6-DoF) object-relative pose measurements. The authors trained a deep learning (DL)-based object pose predictor to directly predict the 6-DoF pose of known objects from RGB images. 

While AI-based methods excel in extracting semantic information from images, e.g., object class and 6-DoF object poses, they still can output erroneous predictions. Especially in the context of EKF-based sensor fusion, it is important to model the measurement covariance correctly, quantifying the sensor observations' uncertainty and ultimately the measurement's importance compared to the predicted state during the update step. In \cite{jantos2025aleatoric}, Jantos et al. extended the DL-based object pose estimator for aleatoric uncertainty quantification, the uncertainty inherent in the input data, and replaced the fixed measurement covariance in an EKF with the dynamically predicted aleatoric uncertainty. They showed that the per-image and -object predicted aleatoric uncertainty captures the error characteristics of the full 6-DoF pose, even indicating ambiguous and difficult scenarios through increased uncertainty values. The latter gives ground for introducing aleatoric uncertainty-based outlier rejection (\texttt{AOR}) of measurements by determining suitable thresholds.

\begin{figure}[t]
    \centering
    \includegraphics[width=0.9\linewidth]{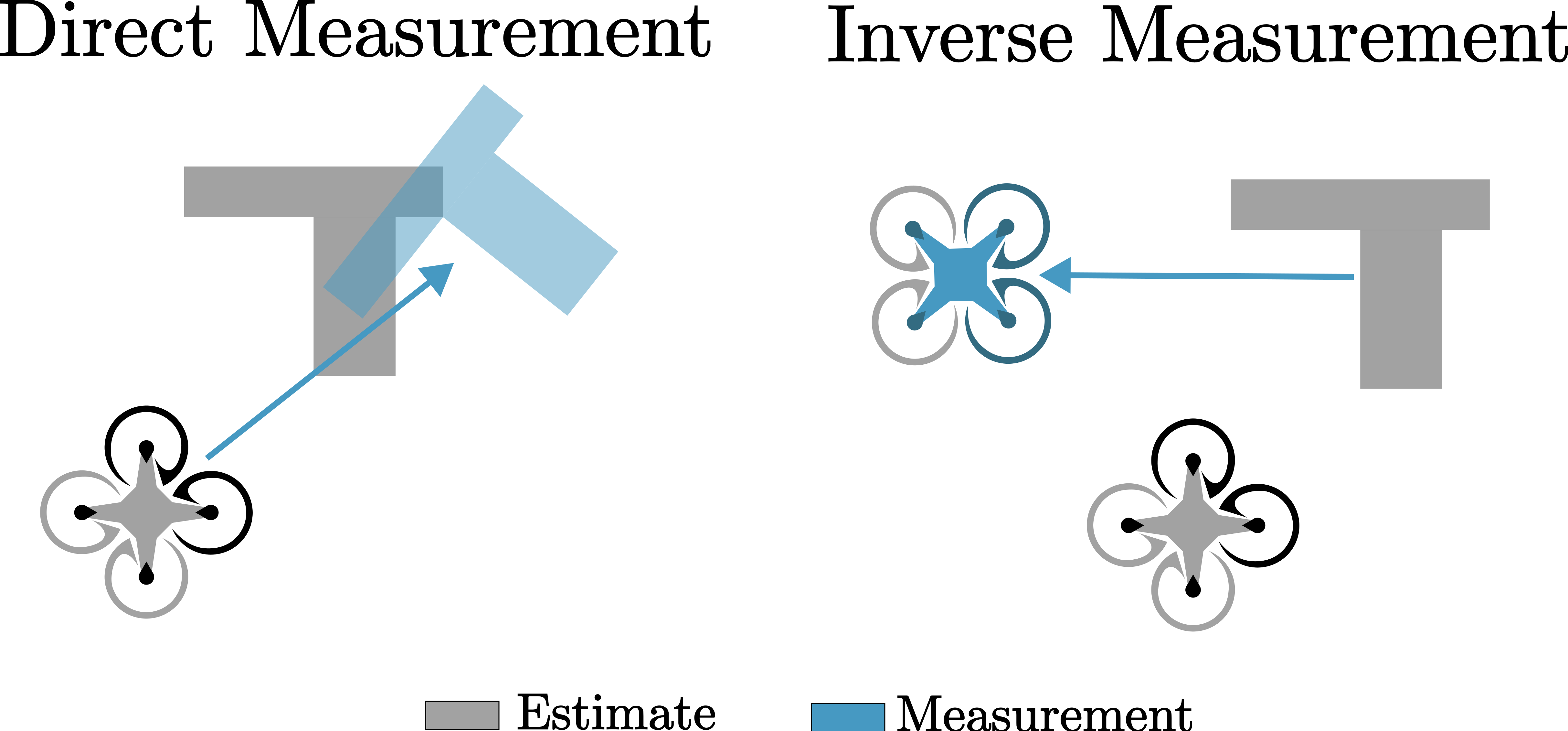}
    \caption{Comparison of the 6-DoF object-relative pose measurements (\texttt{blue}) influence on the state estimation task (\texttt{gray}) for the proposed direct measurement and the inverse measurement \cite{jantos2023aiobjrelstate} approaches. In both cases, the 6-DoF object pose is initially measured with respect to the mobile robot. By inverting the measurement, the relative pose measurement is expressed in the estimated object frame, thus rotation measurement errors lead to discrepancies between the measured and estimated mobile robot 6-DoF pose. In contrast, the proposed direct approach shifts the measurement error towards the object's estimate rather than the mobile robot's state. Moreover, our approach enables the decoupling of the translation and rotation measurement, allowing the rejection of either part.}
    \label{fig:comparison_measurement_definition}
    \vspace{-0.7cm}
\end{figure}

In order to realize their object-relative EKF \cite{jantos2023aiobjrelstate}, it is necessary to invert the DL-based 6-DoF object pose measurements, i.e., instead of using the direct 6-DoF object pose expressed in the camera frame, the EKF measures the camera in the estimated object frame. While this allows for the straightforward derivation of the update Jacobians, it introduces a dependence on the estimated and measured object orientation, as visualized in \cref{fig:comparison_measurement_definition} for a 2D example. Predicting the wrong object orientation also negatively impacts the measured position due to inverting the full 6-DoF pose measurement based on the estimated object frame.
\looseness=-1

In this work, we propose reformulating the EKF to use the direct 6-DoF object pose measurement, thus disentangling the object-relative position and orientation measurement. In addition to improved robustness against incorrect object orientation measurements, often caused by symmetric objects, this gives rise to the possibility of rejecting partial measurements, i.e., only the translation or rotation part of the full 6-DoF pose measurement. In the course of this paper, we will compare classical outlier rejection methods, i.e., $\chi^2$-test, to the \texttt{AOR} approach introduced in \cite{jantos2025aleatoric}, further underlining the benefits of AI-based uncertainty prediction for the state estimation task.
\looseness=-1

\newpage
Our contributions are the following:

\begin{itemize}
    \item Introducing a reformulated filter-based approach for object-relative state estimation that removes the need for measurement inversion.
    \item Decoupling object-relative position and rotation measurements allows for partial measurement rejection and increases robustness towards noisy rotation measurements.
    \item Extending aleatoric uncertainty-based outlier rejection to partial measurements for improved state estimation performance and consistency.
    \item Validating the proposed approach with several experiments and comparisons.
\end{itemize}

The remainder of the paper is organized as follows. We summarize the related work in \cref{sec:related_work}. In \cref{sec:method}, the reformulated EKF for object-relative state estimation, aleatoric uncertainty as dynamic measurement noise covariance, and partial measurement rejection are presented. The experiments and the corresponding results are discussed in \cref{sec:results}. Finally, the paper is concluded in \cref{sec:conclusion}. 

\section{Related Work}
\label{sec:related_work}

Classical approaches for mobile robot state estimation typically combine IMU measurements with sensor modalities such as GNSS \cite{scheiber2023revisiting}, radars \cite{michalczyk2023radar}, or cameras \cite{fornasier2024msceqf} for visual inertial odometry (VIO). However, these approaches are usually unsuitable for object-relative state estimation and navigation as their measurements do not include the necessary semantic information about the objects of interest. Attaching cooperative markers to the objects allows for directly incorporating the 6-DoF object pose into the estimation process \cite{do2023vision,teixeira2018vi,shen2023aeronet}. While straightforward and easy to use, it is often not feasible to equip objects of interest in the wild with markers.

Alternatively, object-relative state estimation can be realized by inferring the 6-DoF object pose from its geometric features. Thomas et al. \cite{thomas2015visual} proposed a method for localization with respect to cylindrical objects, assuming a known radius. Loianno et al. \cite{loianno2018localization} employed a parametric ellipse representation to detect objects in images and estimate their pose from visual attributes such as size, combined with camera parameters. Máthé et al. \cite{mathe2016vision} applied classical techniques to recover full 6-DoF object poses for relative localization, while also exploring the use of machine learning to detect object presence in images. 

Uncertainty quantification enables a better assessment of a deep neural network's (DNN) behavior and gives confidence to its predictions \cite{gawlikowski2023survey}. There are two main sources of uncertainty in DL: aleatoric uncertainty, inherent in the data, and epistemic uncertainty, which is the lack of knowledge in the model. Different approaches exist to model and utilize the uncertainty of 6-DoF pose predictors. Zorina et al. \cite{zorina2025temporally} empirically determine a 6-DoF pose uncertainty to be subsequently used in an object-relative SLAM approach. The covariance matrix assumes a single variance value for the rotational components and a shared variance value for the x and y components. In NVINS \cite{han2024nvins}, a DL-based 6-DoF camera pose predictor is combined with IMU measurements in factor graph optimization. Besides the camera pose, the DNN also predicts its aleatoric and epistemic uncertainty in the form of a 3D Gaussian for the translation and a one-dimensional Langevin distribution of the rotation.

\section{Method}
\label{sec:method}

In this section, we present our approach for 6-DoF object-relative state estimation. After introducing the notation used throughout this paper, the EKF formulation, the update step, and outlier rejection are presented.

\subsection{Notation}
\label{subsec:notation}

Given two coordinate frames \rframe{A} and \rframe{B}, the homogeneous transformation $\vvar{T}[A][B]$ defines frame \rframe{B} with respect to frame \rframe{A}. The transformation consists of the translation $\vvar{p}[A][B] \in \mathbb{R}^3$ and the rotation $\rvar{A}{B} \in SO(3)$. The rotation can also be represented by the quaternion $\vvar{q}[A][B] = [\mathbf{q_v}~q_w]^T = [q_x ~ q_y ~ q_z ~ q_w]^T$. The quaternion multiplication is represented by $\otimes$. An alternative is taking the matrix logarithm of the rotation to get an element of its Lie algebra and the axis-angle representation given by
\begin{align}
    Log(\rvar{}{}) &= \theta [\vvar{u}]_\times \\
    \bm{\vartheta} &= \theta \vvar{u} \in \mathbb{R}^3 \quad ,
\end{align}
with $\theta$ and $\vvar{u}$ being the angle and the axis of rotation, and $[\cdot]_\times$ is the skew-symmetric operator as defined in \cite{sola2017quaternion}. The inverse rotation is expressed by the transposed rotation $\rvar{A}{B} = \rvarT{B}{A}$ or the conjugate quaternion $\vvar{q}[A][B] = \vvari{q}[B][A]$. The inverse position is then given by
\begin{equation}
\label{eq:inverse_pos}
    \vvar{p}[B][A] = - \rvar{B}{A} \vvar{p}[A][B] \quad .
\end{equation}
$\mathbf{I}_N$ and $\mathbf{0}_N$ refer to the identity and the null matrix in $\mathbb{R}^{N\times N}$. 

\subsection{EKF Formulation}
\label{subsec:filter}

\begin{figure}
    \centering
    \includegraphics[width=1.0\linewidth]{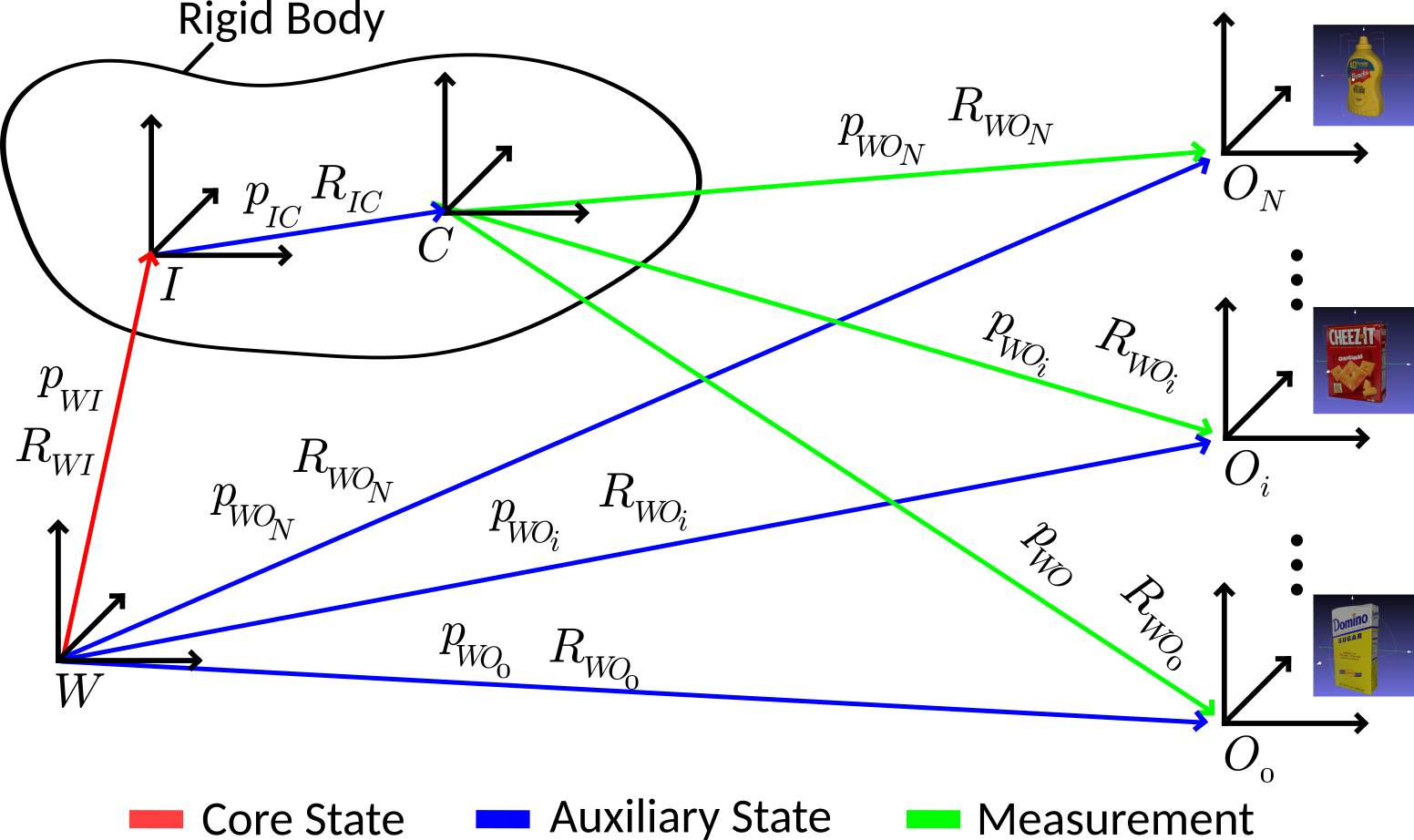}
    \caption{Visualization of the different reference frames. The goal is to estimate the position and orientation of a rigid body (\texttt{red}) consisting of an IMU (\rframe{I}) and camera (\rframe{C}) in a fixed but arbitrary navigation frame (\rframe{W}) with respect to a set of objects of interest (\rframe{O_i}, \texttt{blue}). A DL-based pose predictor provides the object-relative 6-DoF pose measurements to the state estimator (\texttt{green}). The extrinsic calibration between the IMU and the camera is an additional auxiliary state in our formulation.}
    \label{fig:estimated_frames}
    \vspace{-0.7cm}
\end{figure}

In object-relative state estimation, the goal is to estimate the state of a mobile robot (\rframe{I}) in an arbitrary but fixed navigation frame (\rframe{W}) with respect to a set of objects of interest (\rframe{O_i}). The DL-based object pose predictor provides the relative pose measurements between a camera (\rframe{C}) and \rframe{O_i}. The different frames are visualized in \cref{fig:estimated_frames}. Depending on the total number of objects $N$ in a scene, the full state vector $\mathbf{X}$ is then defined as: 
\begin{align}
    \mathbf{X} = [& \vvarT{p}[W][I], \vvarT{v}[W][I], \vvarT{q}[W][I], \vvarT{b}[\omega], \vvarT{b}[a], \nonumber \\
    &\vvarT{p}[I][C], \vvarT{q}[I][C], \vvarT{p}[W][O_0], \vvarT{q}[W][O_0], \dots, \vvarT{p}[W][O_N], \vvarT{q}[W][O_N]]. 
\end{align}
The core states necessary for state propagation are the position $\vvar{p}[W][I]$ of the IMU, its velocity $\vvar{v}[W][I]$ and its orientation $ \vvar{q}[W][I]$ as well as the gyroscopic bias $\vvar{b}[\omega]$ and the accelerometer bias $\vvar{b}[a]$. The pose and velocity dynamics are given as \cite{weiss2011monocular}:
\begin{align}
    \vvard{p}[W][I] &= \vvar{v}[W][I] \\
    \vvard{v}[W][I] &= \rvar{W}{I} (\vvar{a}[m] - \vvar{b}[a] - \vvar{n}[a]) - \vvar{g} \\
    \vvard{q}[W][I] &= \frac{1}{2} \Omega(\bm{\omega} - \vvar{b}[\omega] - \vvar{n}[\omega]) \vvar{q}[W][I],
\end{align}
where ${\mathbf{a}}_{\scriptscriptstyle m}$ is the measured acceleration in \rframe{I}, ${\mathbf{n}}_{\scriptscriptstyle a}$ is the accelerometer noise parameter, $\mathbf{g}$ is the gravity vector in \rframe{W}, ${\mathbf{\omega}}_{\scriptscriptstyle b}$ is the measured angular velocity in \rframe{I}, ${\mathbf{n}}_{\scriptscriptstyle \omega}$ is the gyroscopic noise parameter, and $\Omega(\omega)$ is the quaternion multiplication matrix of $\omega$. The IMU biases are modeled as random walks.

Similar to \cite{jantos2023aiobjrelstate}, we estimate the pose of the objects in the navigation frame ($\vvar{p}[W][O_i], \vvar{q}[W][O_i]$), modeled to stay consistent over time. In contrast and in tone with the direct measurement definition, we estimate the pose of the objects in the navigation frame, rather than the navigation frame with respect to the object frame. The extrinsic calibration between the IMU and the camera is also part of the state and can be estimated. In this work, we assume the calibration to be fixed and known.

The DL-based object pose predictor outputs a 6-DoF pose measurement for each image for each detected and known object. Given the current estimated mobile robot pose ($\vvar{p}[W][I], \vvar{q}[W][I]$) and the measured relative object poses ($\vvarh{p}[C][O_i], \vvarh{q}[C][O_i]$), the projected object frames are calculated with
\begin{align}
\label{eq:proj_obj_frames_pos}
    \vvarh{p}[W][O_i] &= \vvar{p}[W][I] + \rvar{W}{I} (\vvar{p}[I][C] + \rvar{I}{C} \vvarh{p}[C][O_i]) \\
\label{eq:proj_obj_frames_rot}
    \rvarh{W}{O_i} &= \rvar{W}{I} \rvar{I}{C} \rvarh{C}{O_i} \quad .
\end{align}
A Hungarian algorithm, based on the Euclidean and geodesic distance and the object class, matches the predicted object frames to the currently estimated object frames ($\vvar{p}[W][O_i], \vvar{q}[W][O_i]$). When viewed for the first time, \cref{eq:proj_obj_frames_pos,eq:proj_obj_frames_rot} are used to initialize a new object frame in the state $\mathbf{X}$. During the EKF update step, the position and orientation residual for each matched object \rframe{O_i} is calculated independently:
\begin{align}
    \Tilde{\mathbf{z}}_{\vvar{p}[O_i]} &= \hat{\mathbf{z}}_{\vvar{p}[O_i]} - \mathbf{z}_{\vvar{p}[O_i]} \nonumber\\
    &= \vvarh{p}[C][O_i] - (\vvar{p}[C][I] + \rvar{C}{I}(\vvar{p}[I][W] + \rvar{I}{W}\vvar{p}[W][O_i])) \\
    &= \vvarh{p}[C][O_i] - (\rvarT{I}{C} (-\vvar{p}[I][C] + \rvarT{W}{I}(\vvar{p}[W][O_i] - \vvar{p}[W][I]))
\end{align}
\begin{align}
    \Tilde{\mathbf{z}}_{\rvar{O_i}{}} &= 2 \frac{\Tilde{\mathbf{z}}_{\mathbf{q}_{\mathbf{v},\mathsf{O_i}}}}{\Tilde{\mathbf{z}}_{q_{\mathsf{w}, \mathsf{O_i}}}} \\
    \Tilde{\mathbf{z}}_{\vvar{q}[O_i]} &= \mathbf{z}^{-1}_{\vvar{q}[O_i]} \otimes \hat{\mathbf{z}}_{\vvar{q}[O_i]} \nonumber\\
    &= (\vvar{q}[C][I] \otimes \vvar{q}[I][W] \otimes \vvar{q}[W][O])^{-1} \otimes \vvarh{q}[C][O_i] \\
    &= (\vvari{q}[I][C] \otimes \vvari{q}[W][I] \otimes \vvar{q}[W][O])^{-1} \otimes \vvarh{q}[C][O_i] \\
    \label{eq:residual_pose_measurement}
    \Tilde{\mathbf{z}}_{\mathsf{O_i}} &= \begin{bmatrix}
        \Tilde{\mathbf{z}}_{\vvar{p}[O_i]} \\
        \Tilde{\mathbf{z}}_{\vvar{q}[O_i]}
    \end{bmatrix} \quad .
\end{align}
The filter formulation using the inverse 6-DoF object pose measurement \cite{jantos2023aiobjrelstate} suffers from the inverse position measurement being dependent on the currently measured object orientation, see \cref{eq:inverse_pos}. Hence, rotation measurement errors are also expressed in the position measurement, as visualized in \cref{fig:comparison_measurement_definition}.
\looseness=-1

In view of these residuals, it is necessary to determine derivation rules for expressions containing the transpose rotation, see Appendix:
\begin{align}
    \frac{\partial\mathbf{Q}\mathbf{R}^T\mathbf{S}}{\partial\mathbf{R}} = - \mathbf{S}^T \mathbf{R} \qquad \frac{\partial \mathbf{QR}^T\mathbf{v}}{\partial \mathbf{R}} = \mathbf{Q} [\mathbf{R}^T \mathbf{v}]_\mathsf{x} \quad ,
\end{align}
where $\mathbf{Q}, \mathbf{R}, \mathbf{S} \in SO(3)$ and $\mathbf{v} \in \mathbb{R}^3$. Thus, the Jacobians for position $\mathbf{H}_{\mathbf{p}}$ and orientation $\mathbf{H}_{\mathbf{R}}$ for a single relative pose measurement matched to object \rframe{O_i} with respect to the states are \cite{sola2017quaternion}:
\begin{align}
    \label{eq:Hppwi}
    \mathbf{H}_{\mathbf{p},\vvar{p}[W][I]} &= -\rvarT{I}{C}\rvarT{W}{I} \\
    \mathbf{H}_{\mathbf{p}, \rvar{W}{I}} &= \rvarT{I}{C} \left( [\rvarT{W}{I} \vvar{p}[W][O]]_\mathsf{x} - [\rvarT{W}{I} \vvar{p}[W][I]]_\mathsf{x} \right) \\
    \mathbf{H}_{\mathbf{p, \vvar{p}[I][C]}} &= - \rvarT{I}{C} \\
    \mathbf{H}_{\mathbf{p}, \rvar{I}{C}} &= - [\rvarT{I}{C} \vvar{p}[I][C]]_\mathsf{x} \nonumber \\
    & \quad \; + [\rvarT{I}{C} \rvarT{W}{I} \vvar{p}[W][O]]_\mathsf{x} - [\rvarT{I}{C} \rvarT{W}{I} \vvar{p}[W][I]]_\mathsf{x}\\
    \mathbf{H}_{\mathbf{p}, \vvar{p}[W][O_i]} &= \rvarT{I}{C} \rvarT{W}{I} \\
    \mathbf{H}_{\mathbf{p},\rvar{W}{O_i}} &= \mathbf{0}_3 \\
    \label{eq:HRpwi}
    \mathbf{H}_{\mathbf{R},\vvar{p}[W][I]} &= \mathbf{0}_3 \\
    \mathbf{H}_{\mathbf{R}, \rvar{W}{I}} &= -\rvarT{W}{O} \rvar{W}{I} \\
    \mathbf{H}_{\mathbf{R},\vvar{p}[I][C]} &= \mathbf{0}_3 \\
    \mathbf{H}_{\mathbf{R}, \rvar{I}{C}} &= -\rvarT{W}{O} \rvar{W}{I} \rvar{I}{C} \\
    \label{eq:HRpwoi}
    \mathbf{H}_{\mathbf{R},\vvar{p}[W][O_i]} &= \mathbf{0}_3 \\
    \mathbf{H}_{\mathbf{R}, \rvar{W}{O_i}} &= \mathbf{I}_3,
\end{align}

where, e.g. $ H_{\mathbf{p}, {\mathbf{p}}_{\scriptscriptstyle WI}}$ only considers the part of the residual $\Tilde{\mathbf{z}}_{\vvar{p}[O_i]}$ that depends on the state ${\mathbf{p}}_{\scriptscriptstyle WI}$. The rest of the Jacobians are equal to $\textbf{0}_3$. As relative pose measurements for different objects are independent of each other, the Jacobians for the other ($i \neq n$) object-world states, i.e., $\mathbf{H}_{\mathbf{p},\vvar{p}[W][O_n]}$, $\mathbf{H}_{\mathbf{p},\rvar{W}{O_n}}$, $\mathbf{H}_{\mathbf{R},\vvar{p}[W][O_n]}$, $\mathbf{H}_{\mathbf{R},\rvar{W}{O_n}}$ are all equal to $\mathbf{0}_3$. For a single object \rframe{O_i}, the Jacobian is given by stacking the individual components:
\begin{align}
    \mathbf{H}_{\mathbf{p}, \mathsf{O_i}} = [& \mathbf{H}_{\mathbf{p},\vvar{p}[W][I]}, \mathbf{H}_{\mathbf{p},\vvar{v}[W][I]}, \mathbf{H}_{\mathbf{p}, \rvar{W}{I}}, \mathbf{H}_{\mathbf{p}, \vvar{b}[\omega][]}, \mathbf{H}_{\mathbf{p}, \vvar{b}[a][]}, \\
    & \mathbf{H}_{\mathbf{p},\vvar{p}[I][C]}, \mathbf{H}_{\mathbf{p},\rvar{I}{C}}, \mathbf{H}_{\mathbf{p},\vvar{p}[W][O_0]},  \mathbf{H}_{\mathbf{p},\rvar{W}{O_0}}\nonumber \\
    & \dots, \mathbf{H}_{\mathbf{p},\vvar{p}[W][O_N]},  \mathbf{H}_{\mathbf{p},\rvar{W}{O_N}}] \nonumber \\
    \mathbf{H}_{\mathbf{R}, \mathsf{O_i}} = [& \mathbf{H}_{\mathbf{R},\vvar{p}[W][I]}, \mathbf{H}_{\mathbf{R},\vvar{v}[W][I]}, \mathbf{H}_{\mathbf{R}, \rvar{W}{I}}, \mathbf{H}_{\mathbf{R}, \vvar{b}[\omega][]}, \mathbf{H}_{\mathbf{R}, \vvar{b}[a][]}, \\
    & \mathbf{H}_{\mathbf{R},\vvar{p}[I][C]}, \mathbf{H}_{\mathbf{R},\rvar{I}{C}}, \mathbf{H}_{\mathbf{R},\vvar{p}[W][O_0]},  \mathbf{H}_{\mathbf{R},\rvar{W}{O_0}}\nonumber \\
    & \dots, \mathbf{H}_{\mathbf{R},\vvar{p}[W][O_N]},  \mathbf{H}_{\mathbf{R},\rvar{W}{O_N}}] \nonumber
\end{align}
\begin{equation}
\label{eq:jacobians_pose_measurement}
    \mathbf{H}_\mathsf{O_i} = \begin{bmatrix}
        \mathbf{H}_{\mathbf{p}, \mathsf{O_i}} \\
        \mathbf{H}_{\mathbf{R}, \mathsf{O_i}}
    \end{bmatrix} .
\end{equation}

Depending on the current image, as it can capture multiple objects of interest simultaneously, the update is conducted simultaneously. The final residual $\Tilde{\mathbf{z}}$ and observation matrix $\mathbf{H}$ for the state update are determined by vertically stacking the residuals and Jacobians, see \cref{eq:residual_pose_measurement} and \cref{eq:jacobians_pose_measurement}, for each matched object for the current image. As discussed in \cite{jantos2023aiobjrelstate}, one of the object frames ($\vvar{p}[W][O_A], \vvar{q}[W][O_A]$) needs to be fixed, i.e., set its Jacobian $\mathbf{H}_\mathsf{O_A}$ to $\mathbf{0} \in \mathbb{R}^{6 \times 21+3\cdot N}$, to prevent observability issues in pure object-relative state estimation.

\subsection{Outlier Rejection}

Reliably rejecting outlier measurements is crucial for EKF-based state estimation. A commonly used outlier rejection strategy is the $\chi^2$-test. For a single measurement, the $\chi^2$-test checks the statistical plausibility of its innovation covariance $\mathbf{S}$ with
\begin{align}
\label{eq:innovation_cov}
    \mathbf{S} &= \mathbf{H} \mathbf{P} \mathbf{H}^T + \bm{\Sigma} \\
    d^2 &= \Tilde{\mathbf{z}}^T \mathbf{S}^{-1} \Tilde{\mathbf{z}} \quad ,
\end{align}
where $\mathbf{P}$ is the state covariance, $\bm{\Sigma}$ is the measurement noise covariance matrix, and $d^2$ follows a $\chi^2$ distribution with $m$ degrees of freedom. Comparing it to an upper critical value, the statistical consistency of the measurement can be verified. Otherwise, the measurement is rejected. \cref{eq:innovation_cov} indicates that the choice of $\bm{\Sigma}$ directly influences the $\chi^2$-test. If $\bm{\Sigma}$ is chosen too conservatively, the $\chi^2$ will reject too many measurements. On the other hand, large measurement uncertainty values will lead to the inclusion of outlier measurements. Ideally, $\bm{\Sigma}$ should capture the uncertainty of the measurement correctly, a task often not straightforward and requiring time-consuming engineering and tuning efforts. 

We assume no cross-correlations between the individual translation and rotation components, with the latter expressed in the vector space tangent to the measured rotation \cite{sola2017quaternion}. The full measurement noise covariance matrix of the 6-DoF pose measurement expressed in the camera frame is given by

\begin{align}
\label{eq:aleatoric_cov}
    \bm{\Sigma}_{\vvar{p}, \mathsf{CO_i}} &= \begin{pmatrix}
\hat{\sigma}_x^2 & 0 & 0\\
0 & \hat{\sigma}_y^2 & 0 \\
0 & 0 & \hat{\sigma}_z^2
\end{pmatrix} \\
 \bm{\Sigma}_{\bm{\vartheta}, \mathsf{CO_i}} &= \begin{pmatrix}
    \hat{\sigma}_{\vartheta_1}^2 & 0 & 0\\
    0 & \hat{\sigma}_{\vartheta_2}^2 & 0 \\
    0 & 0 & \hat{\sigma}_{\vartheta_3}^2
\end{pmatrix} \\
\label{eq:full_cov_pose_measurement}
\bm{\Sigma}_{\mathsf{CO_i}} &= \begin{pmatrix}
        \bm{\Sigma}_{\vvar{p}, \mathsf{CO_i}} & \textbf{0}_{3\times3} \\
         \textbf{0}_{3\times3} &  \bm{\Sigma}_{\bm{\vartheta}, \mathsf{CO_i}}      
    \end{pmatrix} \quad .
\end{align}

Similar to other measurement modalities, DL-based measurements can suffer from inaccuracies and outliers, due to, e.g., out-of-distribution data or ambiguous viewpoints in the case of 6-DoF object pose estimation. Hence, determining $\bm{\Sigma}$ for a deep neural network requires sophisticated methods. However, a (pre-trained) 6-DoF object pose predictor extended for aleatoric uncertainty prediction can capture its error characteristics \cite{jantos2025aleatoric}, thus serving as a dynamic measurement noise covariance matrix. Moreover, increased aleatoric uncertainty levels indicate ambiguous and challenging situations, ultimately introducing the concept of aleatoric uncertainty-based outlier rejection \texttt{AOR}.

We adopt the aleatoric uncertainty prediction principle to our direct measurement filter formulation. Given an input image, the 6-DoF pose predictor outputs the translation $\vvarh{p}[C][O_i]$, the rotation $\hat{\bm{\vartheta}}_{\mathsf{CO_i}}$ and corresponding aleatoric uncertainties ($\hat{\bm{\Sigma}}_{\vvar{p},\mathsf{CO_i}}, \hat{\bm{\Sigma}}_{\bm{\vartheta},\mathsf{CO_i}}$), modeled to represent Gaussian distributions with $\mathcal{N}(\vvarh{p}[C][O_i], \hat{\bm{\Sigma}}_{\vvar{p},\mathsf{CO_i}})$ and $\mathcal{N}(\hat{\bm{\vartheta}}_{\mathsf{CO_i}}, \hat{\bm{\Sigma}}_{\bm{\vartheta}, \mathsf{CO_i}})$. The predicted aleatoric measurement covariance matrices can be used as our state estimator's measurement noise covariance matrix. 

As the measurement noise covariance is expressed in the camera's frame and due to our direct filter formulation, the need for inversion of the measurement covariance matrix is eliminated. The inverse measurement covariance matrix depends on the predicted rotation:
\begin{equation}
\label{eq:inverse_cov}
 \bm{\Sigma}_{\mathbf{y}, \mathsf{O_iC}} = \rvarh{O_i}{C} \bm{\Sigma}_{\mathbf{y}, \mathsf{CO_i}} \rvarhT{O_i}{C}, \quad \mathbf{y} \in \{\mathbf{p}, \bm{\vartheta}\} .
\end{equation}
Similar to the inversion of the position measurement, erroneous rotation measurements negatively impact the EKF steps involving the measurement noise covariance matrix. Once again, our direct filter formulation enables the individual consideration of the position and rotation measurement, also for outlier measurement rejection. With partial measurement rejection in place, the EKF can incorporate more information into the estimation process.   

We expand \texttt{AOR} for partial measurement rejection by comparing the positional and rotational uncertainties to a fixed position and rotation uncertainty threshold. If the predicted uncertainty of a single component is above the threshold, the respective position or rotation measurement is rejected. Independent of the outlier rejection method, the rejection of the full or partial measurement requires the appropriate removal and restacking of the residual, Jacobians, and measurement noise covariance matrix, see \cref{eq:residual_pose_measurement,eq:jacobians_pose_measurement,eq:full_cov_pose_measurement}.

\section{Experiments \& Results}
\label{sec:results}

In this section, the conducted experiments are presented and the corresponding results are discussed. Before introducing the 6-DoF pose prediction framework, we go into the dataset used for the experiments. Afterwards, we show the strengths of our filter formulation in comparison to the EKF presented in \cite{jantos2023aiobjrelstate}. Finally, the influence of aleatoric uncertainty and partial measurement rejection for the object-relative state estimation task is highlighted.   

\subsection{Dataset}

The experiments focus on a subset of the commonly used YCB-V object set \cite{xiang2018posecnn}. This subset covers a wide range of object attributes, ranging from well-textured (cracker box, sugar box, power drill) to having ambiguous viewpoints (mustard bottle, bleach cleanser) to being almost symmetrical (scissors, mug). This work focuses on synthetic data as it offers controlled conditions with perfect annotation to perform meaningful analysis of the proposed approach.
\looseness=-1

\begin{figure}
    \centering
    \includegraphics[width=0.44\linewidth]{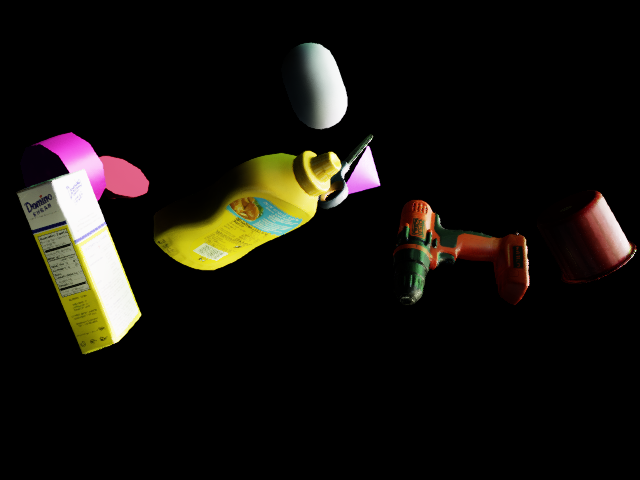}
    \includegraphics[width=0.44\linewidth]{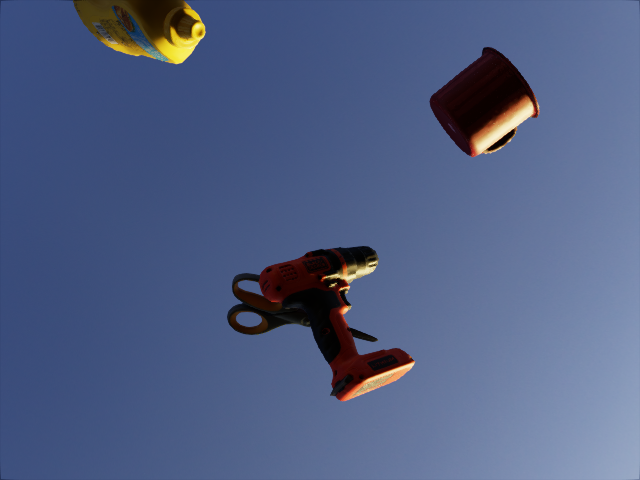}
    \caption{\texttt{Left}: Example of a synthetic training image containing the subset of the YCB-V objects and distractor objects. \texttt{Right}: Example image from an evaluation trajectory with a challenging object constellation and occlusion.}
    \label{fig:synt_example}
\end{figure}

NVIDIA Omniverse IsaacSim allows for the efficient generation and annotation of photorealistic RGB images, given the already included 3D models of the objects. To train and validate the 6-DoF object pose estimator, we generate 100,000 and 3000 synthetic images, respectively. To evaluate the object-relative state estimator, we generate ten diverse physically feasible trajectories with different objects and constellations, and varying distances to the objects. The data includes synthetic IMU data at a rate of 200 Hz and corresponding RGB images recorded with 20 FPS. Example synthetic images are shown in \cref{fig:synt_example}.

\begin{figure}[t]
    \centering
    \includegraphics[width=1.0\linewidth]{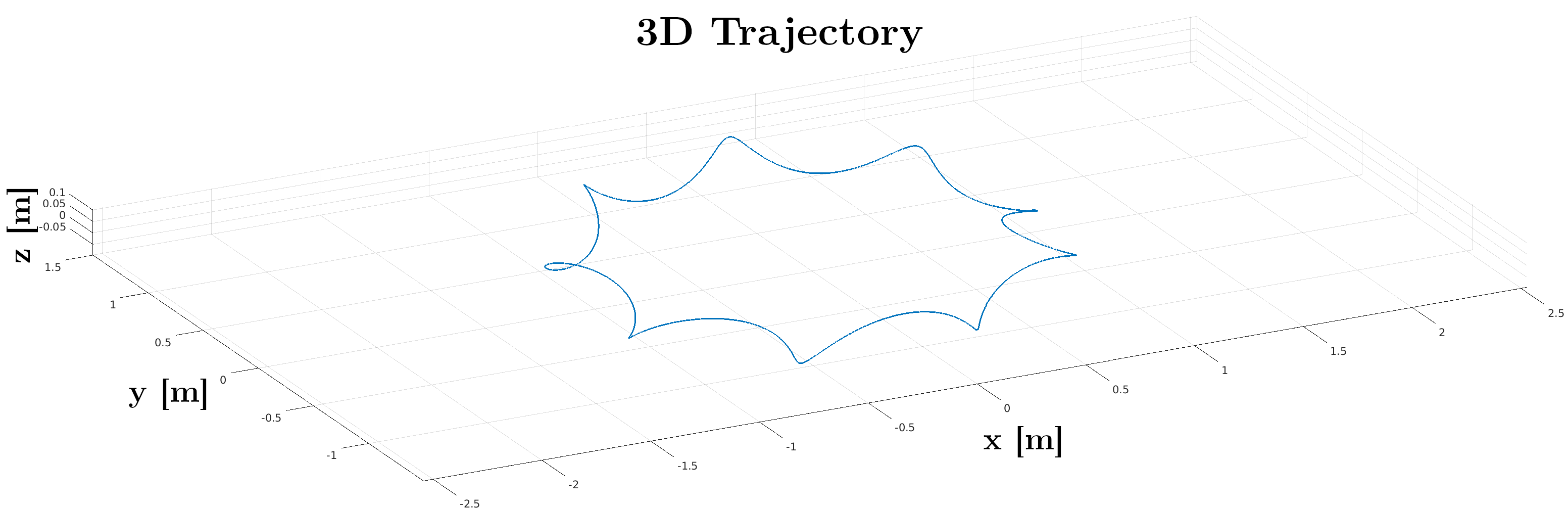}
    \caption{Example trajectory used for evaluation with varying distances and viewing angles of the objects.}
    \label{fig:synt_traj}
\end{figure}

\subsection{DL 6-DoF Object Pose Prediction Framework}
\label{subsec:pose_estimator}

We follow previous work on AI-based object-relative state estimation \cite{jantos2025aleatoric} and use PoET \cite{jantos2023poet} for 6-DoF object pose and aleatoric uncertainty prediction. The object detection backbone is a Scaled-YOLOv4 \cite{wang2021scaledyolo}, and the transformer consists of five encoder and decoder layers and 16 attention heads. The translation, rotation, and aleatoric uncertainty heads are simple multi-layer perceptrons with three layers and an output dimension of three. We first train PoET for 50 epochs before calibrating the aleatoric uncertainty heads for 10 additional epochs, while freezing the remainder of the network.
\looseness=-1
\begin{figure*}[t]
    \centering
    \includegraphics[width=0.49\textwidth]{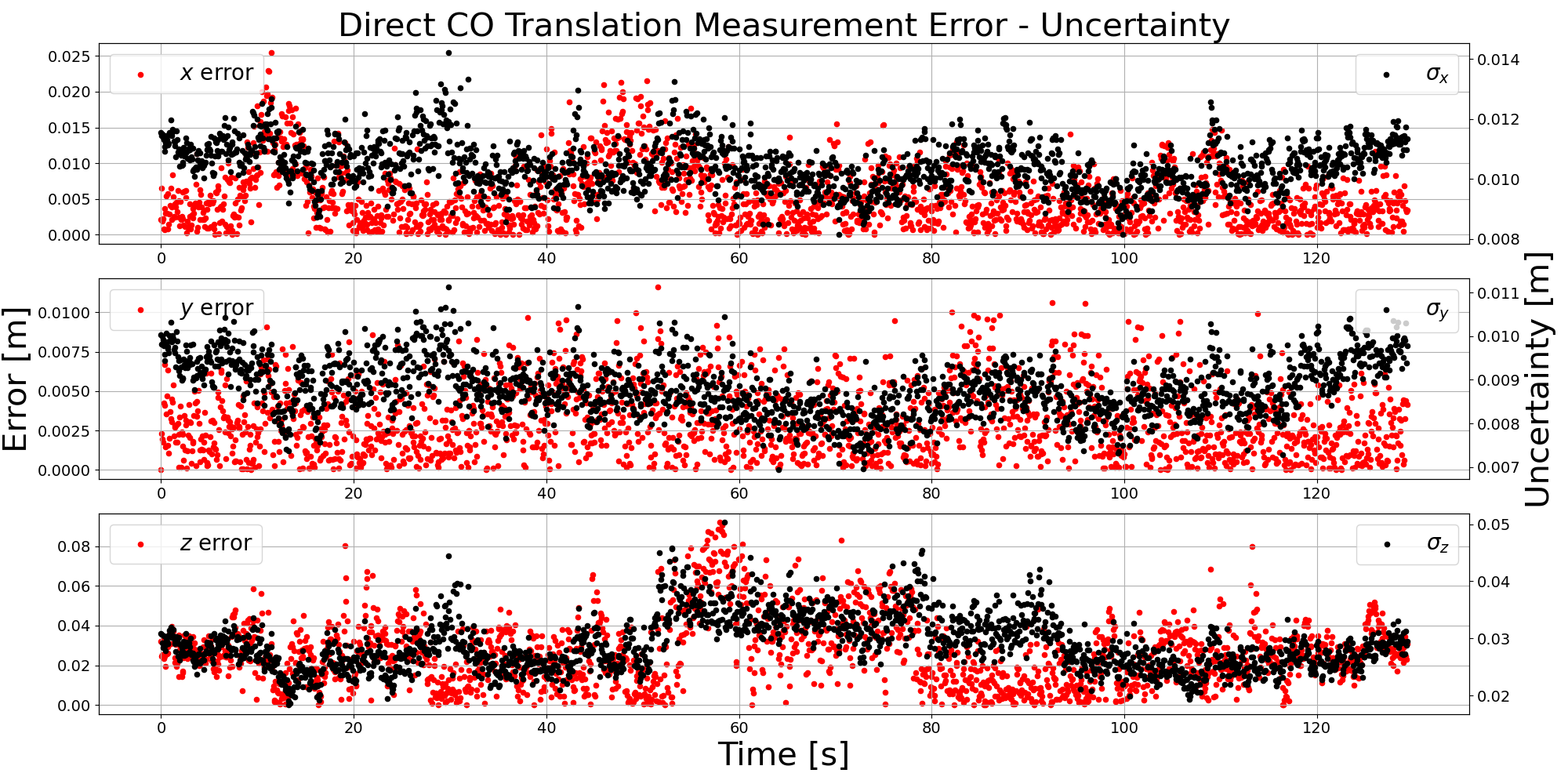}
    \includegraphics[width=0.49\textwidth]{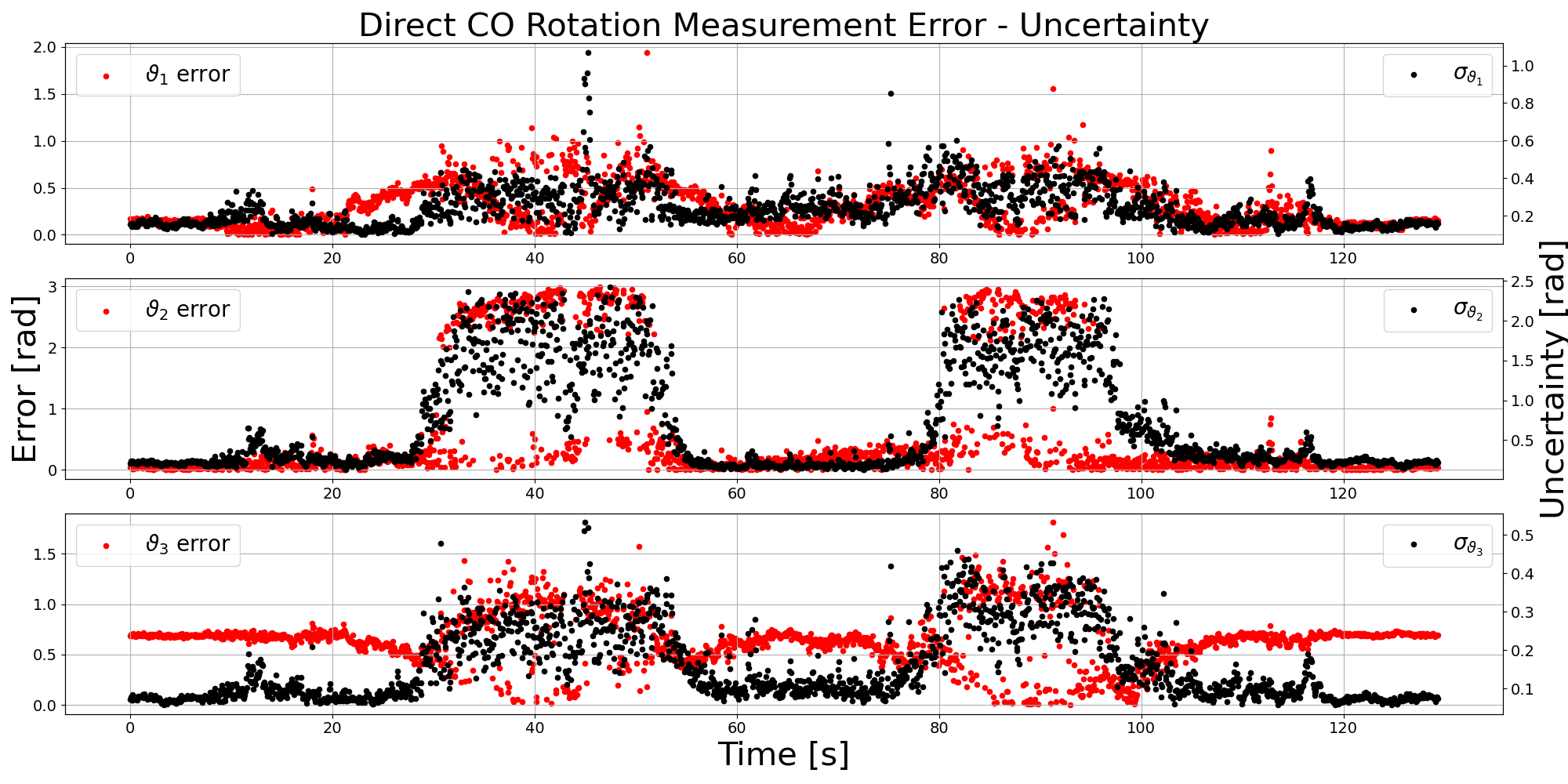}
    \caption{\textbf{Direct Measurement (ours)}: Comparison of the absolute translation error (\texttt{left}, \texttt{red}) and rotation error (\texttt{right}, \texttt{red}) to the estimated aleatoric uncertainty (\texttt{black}) across the whole trajectory for the mug. Note the different scales in the plots' axes and across the plots.}
    \label{fig:direct_measurement_mug}
    \vspace{-0.2cm}
\end{figure*}

\begin{figure*}[t]
    \centering
    \includegraphics[width=0.49\textwidth]{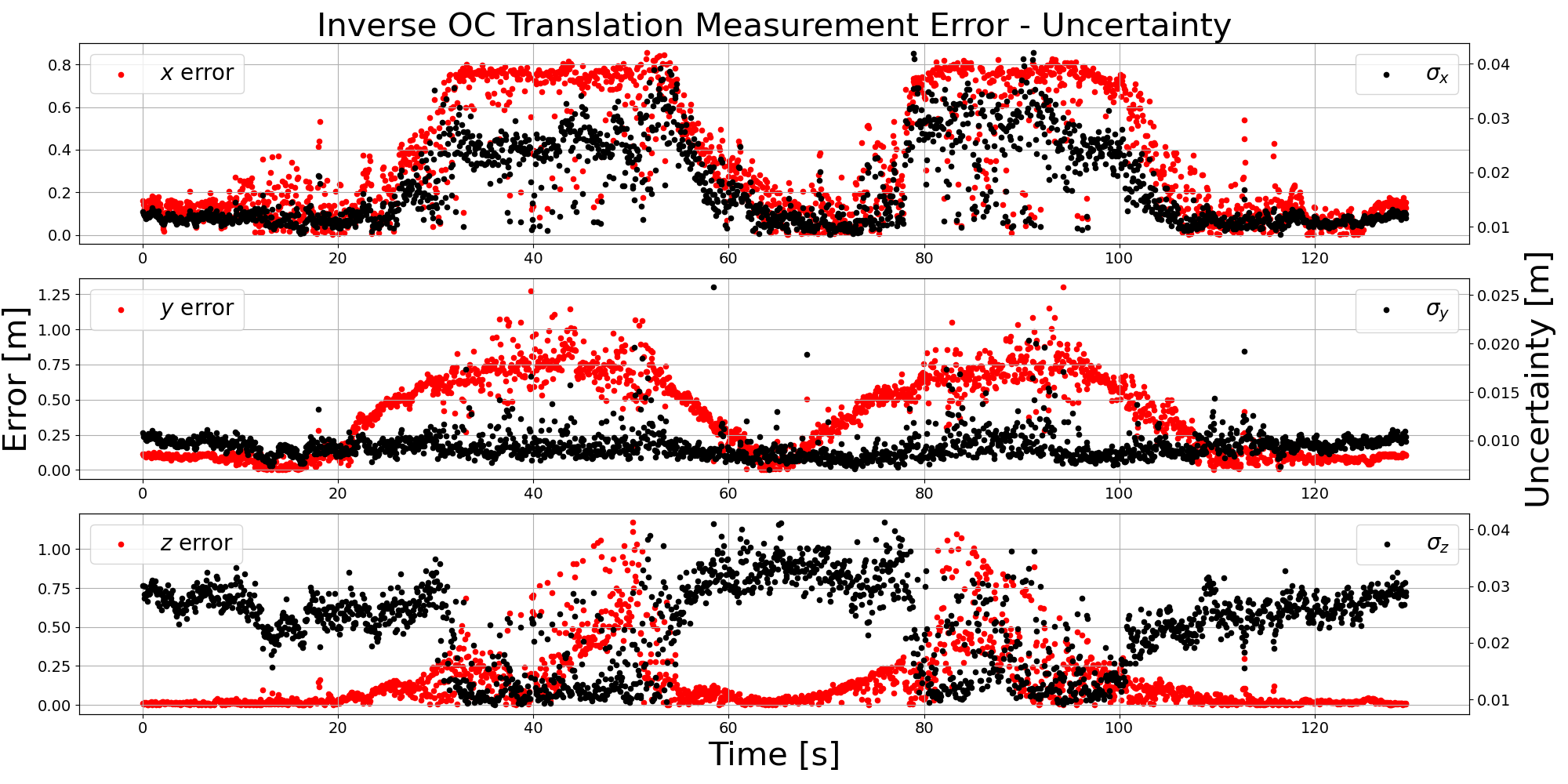}
    \includegraphics[width=0.49\textwidth]{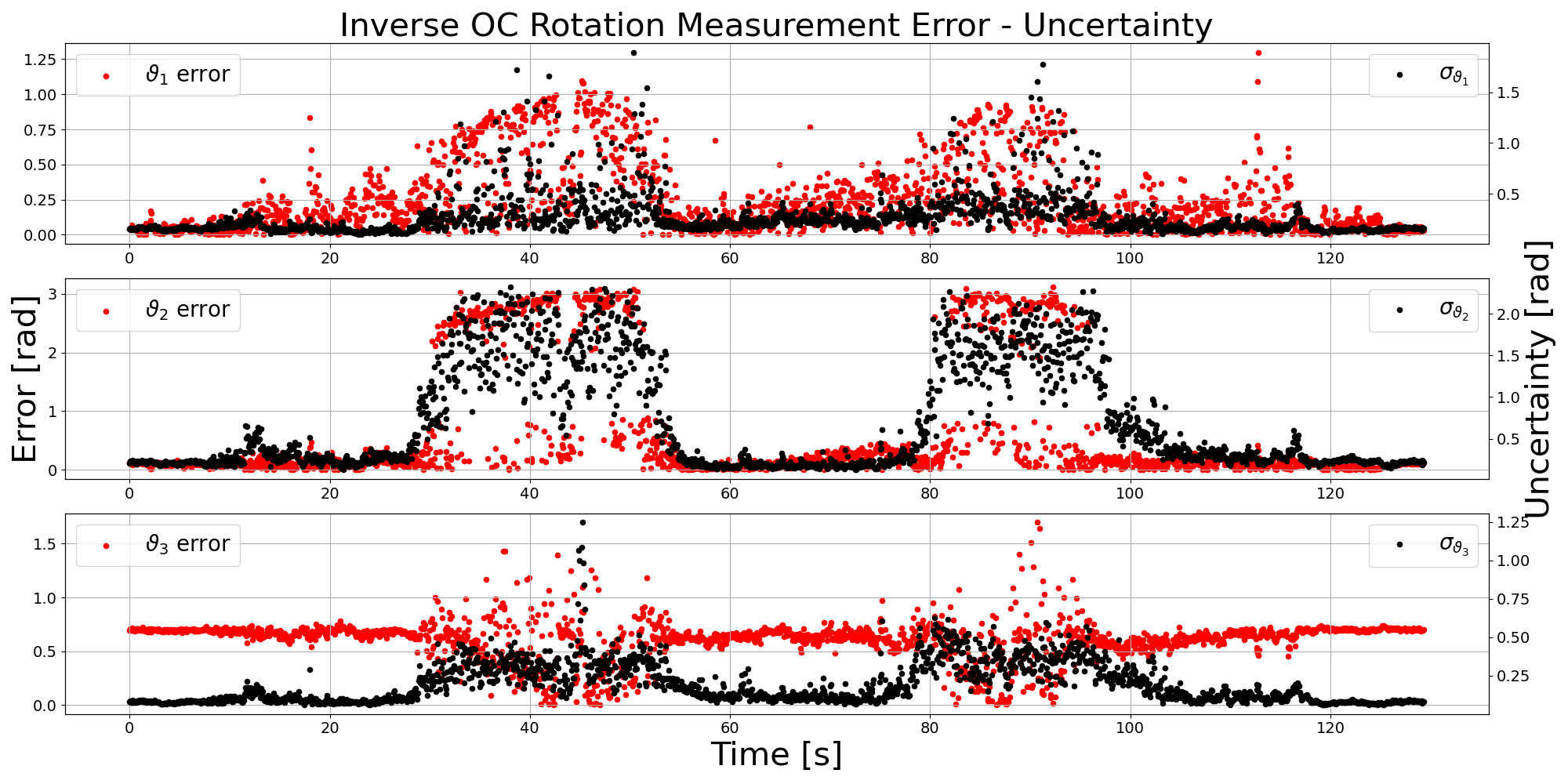}
    \caption{\textbf{Inverse Measurement (\cite{jantos2023aiobjrelstate})}: Comparison of the absolute translation error (\texttt{left}, \texttt{red}) and rotation error (\texttt{right}, \texttt{red}) to the estimated aleatoric uncertainty (\texttt{black}) across the whole trajectory for the mug. Note the different scales in the plots' axes and across the plots.}
    \label{fig:inverse_measurement_mug}
    \vspace{-0.5cm}
\end{figure*}

\subsection{Filter Comparison}
\label{subsec:filter_comparison}

In order to compare the proposed object-relative filter formulation to the inverse formulation presented in \cite{jantos2023aiobjrelstate}, we implement both EKFs using MaRS \cite{brommer2020mars} and report their position root mean square error (RMSE) in meters. As pointed out in \cref{subsec:filter}, the main benefit of our approach is the decoupling of position and rotation of the relative 6-DoF pose measurement. Moreover, by not relying on inverting the full measurement, the position measurement is not negatively impacted by wrong rotation measurements, as shown in \cref{fig:comparison_measurement_definition}.
\looseness=-1

For an unbiased comparison, we generate perfect synthetic IMU and relative pose measurement data and perturb the translation and rotation with Gaussian noise with different standard deviations. The following noise values are chosen:
\begin{align*}
    \sigma_x &= \sigma_y = \sigma_z \in \{0.01, 0.05, 0.1, 0.2, 0.3\} [\text{m}]\\
    \sigma_{\vartheta_1} &= \sigma_{\vartheta_2} = \sigma_{\vartheta_3} \in \{0.0175, 0.0875, 0.175, 0.35\} [\text{rad}] \quad .
\end{align*}

For each noise value pair, 100 Monte Carlo simulations are performed. Please note that the EKF is provided with a measurement noise covariance corresponding to the perturbation noise. The mean position RMSE and standard deviation are reported in \cref{tab:direct_noise_rmse_pos,tab:inverse_noise_rmse_pos}.

\begin{table}
    \caption{Noise Analysis for the direct measurement filter formulation. We report the RMSE in [m]}

\centering
\scalebox{0.9}{
\begin{tabular}{c|c|c|c|c}
$\bm{\Sigma}_{\vvar{p}} / \bm{\Sigma}_{\bm{\vartheta}}$ & 1$^\circ$ & 5$^\circ$ & 10$^\circ$ & 20$^\circ$\tabularnewline
\hline 
\hline 
1cm & 0.073 \textpm{} 0.030 & 0.285 \textpm{} 0.123 & 0.377 \textpm{} 0.183 & 0.370 \textpm{} 0.204\tabularnewline
5cm & 0.111 \textpm{} 0.033 & 0.327 \textpm{} 0.151 & 0.475 \textpm{} 0.283 & 0.719 \textpm{} 0.422\tabularnewline
10cm & 0.179 \textpm{} 0.061 & 0.343 \textpm{} 0.115 & 0.481 \textpm{} 0.270 & 0.789 \textpm{} 0.529\tabularnewline
30cm & 0.477 \textpm{} 0.191 & 0.622 \textpm{} 0.198 & 0.737 \textpm{} 0.267 & 1.075 \textpm{} 0.496\tabularnewline
\end{tabular}
}
    \label{tab:direct_noise_rmse_pos}
    
\end{table}

\begin{table}
   \caption{Noise Analysis for the inverse measurement filter formulation. We report the RMSE in [m]}

\centering
\scalebox{0.9}{
\begin{tabular}{c|c|c|c|c}
$\bm{\Sigma}_{\vvar{p}} / \bm{\Sigma}_{\bm{\vartheta}}$ & 1$^\circ$ & 5$^\circ$ & 10$^\circ$ & 20$^\circ$\tabularnewline
\hline 
\hline 
1cm & 0.082 \textpm{} 0.028 & 0.944 \textpm{} 0.893 & 2.197 \textpm{} 0.419 & 2.538 \textpm{} 0.449\tabularnewline
5cm & 0.113 \textpm{} 0.033 & 0.420 \textpm{} 0.163 & 1.686 \textpm{} 0.886 & 2.480 \textpm{} 0.449\tabularnewline
10cm & 0.180 \textpm{} 0.061 & 0.410 \textpm{} 0.130 & 1.063 \textpm{} 0.637 & 2.290 \textpm{} 0.550\tabularnewline
30cm & 0.477 \textpm{} 0.191 & 0.630 \textpm{} 0.190 & 0.979 \textpm{} 0.401 & 1.933 \textpm{} 0.578\tabularnewline
\end{tabular}
}
    \label{tab:inverse_noise_rmse_pos}
\end{table}

In the presence of minimal rotational perturbation, i.e., 1$^\circ$, both filter formulations achieve an almost identical performance independent of the translation perturbation. Given perfect measurement data with no perturbation, both filters achieve the same RMSE of 1.3cm. Increasing the rotation perturbation gradually worsens the performance of both approaches. However, the inverse filter formulation is more affected. This highlights that removing the inversion of the 6-DoF drastically benefits the state estimation performance as it reduces the influence of the rotation component. Interestingly, while increased translation measurement perturbation worsens our state estimator's performance, it has a soothing effect for the filter using the inverse measurement. This is mainly due to the increased measurement noise, which smooths out the measurements. The influence of the inversion of the measurement and its noise covariance matrix is further visualized in \cref{fig:direct_measurement_mug,fig:inverse_measurement_mug}. The plots show the predicted 6-DoF pose measurements and corresponding aleatoric uncertainties for the mug. When viewed from an ambiguous viewpoint, i.e., the occlusion of the handle ($\sim$40s \& $\sim$90s), the rotational error and the aleatoric uncertainty are increased. While the direct translation measurement is unaffected, the rotational errors are additionally reflected in the inverse translation measurement. Besides the inverted position covariance matrix not matching the error characteristics, the numeric values do not adequately capture the error. This leads to either including outlier measurements with high confidence into the state estimation process or to the rejection of originally accurate measurements. Inverting the measurement noise covariance matrix, according to \cref{eq:inverse_cov}, will introduce cross-covariances between the individual position and rotation components, but not between the position and rotation measurements.
\looseness=-1

\subsection{Partial Measurement Rejection}

To highlight the benefit of aleatoric uncertainty and partial measurement rejection for object-relative state estimation, we compare different combinations of measurement noise covariances and outlier rejection methods. In each case, the predicted relative pose measurements of PoET are used. The performance of the state estimator is measured in terms of the RMSE in position and orientation, the maximum position error, and the average normalized estimation error squared (ANEES) for the position and orientation. For better readability, the ANEES is further normalized by the degrees of freedom of the variable, i.e., optimal values are close to 1.
\looseness=-1

First, we use a fixed measurement noise covariance matrix (\texttt{F}). It is determined by calculating the average translation and rotation error across all objects for the validation dataset, resulting in the following measurement noise values:

\begin{align*}
    \sigma_x &= \sigma_y = \sigma_z = 0.04 [\text{m}]\\
    \sigma_{\vartheta_1} &= \sigma_{\vartheta_2} = \sigma_{\vartheta_3} = 0.628 [\text{rad}] \quad .
\end{align*}

Second, we use the per-image and -object predicted aleatoric uncertainty as dynamic measurement noise covariance (\texttt{U}). In terms of outlier rejection, we employ the $\chi^2$-test for statistical outlier rejection, aleatoric uncertainty-based outlier rejection (\texttt{AOR}), and their counterparts for partial measurement rejection ($\chi^2$\texttt{P}, \texttt{AORP}). When determining suitable thresholds for \texttt{AOR}(\texttt{P}), several aspects need to be taken into consideration, such as applicability to different objects and scenarios, and finding the balance between incorporating outlier measurements and rejecting too many measurements, favoring dead reckoning. The experiments conducted in \cref{subsec:filter_comparison} show that measurements with a perturbation noise of 0.1m and 0.175rad already lead to a deteriorating performance. Hence, we choose these values as our thresholds for \texttt{AORP}. As \texttt{AOR} rejects the whole measurement, we choose more conservative threshold values, i.e., 0.15m and 0.35rad, to prevent the rejection of too many measurements and dead reckoning. Please note that the combination of \texttt{F} and \texttt{AOR}(\texttt{P}) is not investigated as thresholding would either accept or reject all measurements. The results for the ten trajectories and the mean are reported in \cref{tab:ycbv_rmse_pos,tab:ycbv_rmse_ori,tab:ycbv_max,tab:ycbv_nees_pos,tab:ycbv_nees_ori}. Bold values indicate the best-performing uncertainty and outlier rejection combination, while a dash (---) indicates that the state estimator diverged for this particular trajectory. These runs are excluded from the mean calculation, skewing the results for the specific combination, and are marked in italic.
\looseness=-1

\begin{table}
    \caption{RMSE Position {[}m{]}}

\centering
\scalebox{0.85}{
\begin{tabular}{c|c|c|c|c|c|c|c|c}
& \texttt{F} & + $\chi^2$ & + $\chi^2$\texttt{P} & \texttt{U} & + $\chi^2$ & + $\chi^2$\texttt{P} & + \texttt{AOR} & + \texttt{AORP}\tabularnewline
\hline 
\hline 
1 & --- & 0.082 & 0.076 & 0.056 & 0.058 & 0.058 & \textbf{0.051} & 0.061\tabularnewline
2 & 0.747 & 0.116 & 0.111 & 0.295 & 0.091 & 0.088 & --- & \textbf{0.079}\tabularnewline
3 & --- & 0.118 & 0.112 & --- & \textbf{0.093} & --- & 0.141 & 0.098\tabularnewline
4 & 0.103 & \textbf{0.086} & 0.087 & 0.109 & 0.108 & 0.107 & 0.115 & 0.104\tabularnewline
5 & 0.553 & 0.119 & 0.123 & 0.281 & 0.113 & 0.109 & 0.154 & \textbf{0.090}\tabularnewline
6 & 0.741 & 0.195 & 0.192 & 0.595 & 0.204 & 0.207 & --- & \textbf{0.114}\tabularnewline
7 & --- & 0.130 & 0.133 & 0.175 & 0.166 & 0.158 & 0.161 & \textbf{0.082}\tabularnewline
8 & --- & 0.132 & 0.140 & --- & 0.137 & 0.140 & 0.159 & \textbf{0.123}\tabularnewline
9 & 0.728 & 0.726 & 0.582 & 0.622 & 0.444 & \textbf{0.101} & --- & 0.123\tabularnewline
10 & 0.770 & 0.230 & 0.200 & --- & 0.119 & \textbf{0.116} & 0.281 & 0.195\tabularnewline
\hline 
Mean & \textit{0.607} & 0.193 & 0.176 & \textit{0.305} & 0.153 & \textit{0.120} & \textit{0.152} & \textbf{0.107}\tabularnewline
\hline 
\end{tabular}
}
    \label{tab:ycbv_rmse_pos}
    \caption{RMSE Orientation [°]}

\centering

\scalebox{0.85}{
\begin{tabular}{c|c|c|c|c|c|c|c|c}
& \texttt{F} & + $\chi^2$ & + $\chi^2$\texttt{P} & \texttt{U} & + $\chi^2$ & + $\chi^2$\texttt{P} & + \texttt{AOR} & + \texttt{AORP}\tabularnewline
\hline 
\hline 
1 & --- & 4.30 & 4.06 & 3.28 & 3.29 & 3.23 & \textbf{3.00} & 3.35\tabularnewline
2 & 44.28 & 6.4 & 6.17 & 16.43 & 5.02 & 4.83 & --- & \textbf{4.21}\tabularnewline
3 & --- & 6.69 & 6.26 & --- & \textbf{5.30} & --- & 7.97 & 5.53\tabularnewline
4 & 5.64 & \textbf{4.59} & 4.64 & 6.08 & 5.99 & 5.97 & 6.41 & 5.77\tabularnewline
5 & 29.53 & 5.84 & 5.96 & 14.12 & 5.02 & 4.84 & 7.04 & \textbf{3.93}\tabularnewline
6 & 38.32 & 9.37 & 9.19 & 30.59 & 9.84 & 9.98 & --- & \textbf{5.34}\tabularnewline
7 & --- & 6.61 & 6.81 & 10.08 & 9.1 & 8.69 & 8.88 & \textbf{4.55}\tabularnewline
8 & --- & 5.33 & 5.75 & --- & 5.63 & 5.74 & 7.27 & \textbf{5.11}\tabularnewline
9 & 43.63 & 42.49 & 34.04 & 36.69 & 25.70 & \textbf{5.17} & --- & 6.32\tabularnewline
10 & 49.43 & 13.07 & 11.16 & --- & 6.28 & \textbf{6.09} & 15.49 & 10.98\tabularnewline
\hline 
Mean & \textit{35.1}4 & 10.47 & 9.40 & \textit{16.75} & 8.12 & \textit{6.06} & \textit{8.01} & \textbf{5.51}\tabularnewline
\hline 
\end{tabular}
}
    \label{tab:ycbv_rmse_ori}
    \caption{Maximum Position Error {[}m{]}}

\centering{}%

\scalebox{0.85}{
\begin{tabular}{c|c|c|c|c|c|c|c|c}
& \texttt{F} & + $\chi^2$ & + $\chi^2$\texttt{P} & \texttt{U} & + $\chi^2$ & + $\chi^2$\texttt{P} & + \texttt{AOR} & + \texttt{AORP}\tabularnewline
\hline 
\hline 
1 & --- & 0.242 & 0.225 & 0.339 & 0.140 & 0.139 & 0.148 & \textbf{0.133}\tabularnewline
2 & 2.677 & 0.253 & 0.242 & 1.412 & 0.224 & 0.218 & --- & \textbf{0.192}\tabularnewline
3 & --- & 0.245 & 0.269 & --- & 0.217 & --- & 0.640 & \textbf{0.241}\tabularnewline
4 & 0.297 & \textbf{0.219} & 0.223 & 0.240 & 0.239 & 0.237 & 0.277 & 0.271\tabularnewline
5 & 1.827 & 0.214 & 0.215 & 0.712 & 0.217 & 0.212 & 0.337 & \textbf{0.190}\tabularnewline
6 & 1.846 & 0.527 & 0.505 & 1.87 & 0.406 & 0.414 & --- & \textbf{0.257}\tabularnewline
7 & --- & 0.301 & 0.316 & 0.728 & 0.305 & 0.310 & 0.277 & \textbf{0.169}\tabularnewline
8 & --- & 0.334 & 0.370 & --- & 0.297 & 0.306 & 0.293 & \textbf{0.268}\tabularnewline
9 & 2.231 & 1.309 & 1.184 & 1.253 & 0.743 & 0.317 & --- & \textbf{0.262}\tabularnewline
10 & 2.712 & 0.525 & 0.591 & --- & 0.372 & \textbf{0.273} & 0.938 & 0.442\tabularnewline
\hline 
Mean & \textit{1.932} & 0.417 & 0.414 & \textit{0.936} & 0.316 & \textit{0.270} & \textit{0.416} & \textbf{0.242}\tabularnewline
\hline 
\end{tabular}
}
    \label{tab:ycbv_max}
    \caption{Normalized ANEES for Position}

\centering
\scalebox{0.85}{
\begin{tabular}{c|c|c|c|c|c|c|c|c}
& \texttt{F} & + $\chi^2$ & + $\chi^2$\texttt{P} & \texttt{U} & + $\chi^2$ & + $\chi^2$\texttt{P} & + \texttt{AOR} & + \texttt{AORP}\tabularnewline
\hline 
\hline 
1 & --- & 3.16 & \textbf{2.84} & 6.52 & 6.54 & 6.60 & 4.53 & 7.04\tabularnewline
2 & 174.3 & 2.35 & \textbf{2.23} & 51.00 & 4.11 & 3.93 & --- & 3.64\tabularnewline
3 & --- & \textbf{1.83} & 2.42 & --- & 4.00 & --- & 9.77 & 4.21\tabularnewline
4 & 2.28 & \textbf{1.57} & 1.62 & 6.61 & 6.42 & 6.38 & 6.54 & 2.80\tabularnewline
5 & 55.00 & 2.62 & 2.97 & 22.31 & 4.16 & 4.12 & 2.63 & \textbf{2.14}\tabularnewline
6 & 96.52 & 2.78 & 2.71 & 151.0 & 7.41 & 7.71 & --- & \textbf{1.01}\tabularnewline
7 & --- & \textbf{4.84} & 5.20 & 33.65 & 31.57 & 29.27 & 31.60 & 6.44\tabularnewline
8 & --- & 4.67 & 5.29 & --- & 6.21 & 6.68 & 4.72 & \textbf{4.46}\tabularnewline
9 & 346.1 & 214.7 & 126.9 & 402.9 & 125.9 & \textbf{7.55} & --- & 10.90\tabularnewline
10 & 325.8 & 14.75 & 13.28 & --- & \textbf{6.43} & 7.64 & 8.26 & 17.13\tabularnewline
\hline 
Mean & \textit{166.7} & 25.33 & 16.54 & \textit{96.28} & 20.28 & \textit{8.88} & \textit{6.54} & \textbf{5.98}\tabularnewline
\hline 
\end{tabular}
}
    \label{tab:ycbv_nees_pos}
    \caption{Normalized ANEES for Orientation}

\centering
\scalebox{0.85}{
\begin{tabular}{c|c|c|c|c|c|c|c|c}
& \texttt{F} & + $\chi^2$ & + $\chi^2$\texttt{P} & \texttt{U} & + $\chi^2$ & + $\chi^2$\texttt{P} & + \texttt{AOR} & + \texttt{AORP}\tabularnewline
\hline 
\hline 
1 & --- & 2.18 & \textbf{1.91} & 6.33 & 5.73 & 5.65 & 4.29 & 5.74\tabularnewline
2 & 110.0 & 1.51 & \textbf{1.41} & 54.33 & 3.43 & 3.19 & --- & 2.27\tabularnewline
3 & --- & \textbf{1.27} & 1.61 & --- & 3.47 & --- & 7.54 & 3.24\tabularnewline
4 & 1.27 & 0.79 & \textbf{0.81} & 5.11 & 4.94 & 4.93 & 5.03 & 2.03\tabularnewline
5 & 56.67 & 1.23 & 1.29 & 23.49 & 2.41 & 2.30 & 1.95 & \textbf{1.08}\tabularnewline
6 & 43.12 & 1.49 & \textbf{1.43} & 258.7 & 4.56 & 4.64 & --- & 0.34\tabularnewline
7 & --- & \textbf{2.83} & 3.14 & 28.03 & 26.35 & 24.07 & 26.27 & 5.26\tabularnewline
8 & --- & 0.70 & 0.83 & --- & 2.78 & 3.00 & 3.15 & \textbf{0.87}\tabularnewline
9 & 317.3 & 171.0 & 104.54 & 433.1 & 111.4 & \textbf{5.30} & --- & 7.51\tabularnewline
10 & 1265 & 10.60 & 8.56 & 495.2 & \textbf{4.35} & 5.06 & 11.93 & 12.01\tabularnewline
\hline 
Mean & \textit{299.0} & 19.36 & 12.55 & \textit{163.0} & 16.94 & \textit{6.46} & \textit{8.59} & \textbf{4.04}\tabularnewline
\hline 
\end{tabular}
}
    \label{tab:ycbv_nees_ori}
\end{table}

On average, using aleatoric uncertainty (\texttt{U}) outperforms its fixed measurement noise covariance matrix (\texttt{F}) counterparts. Besides improved estimation error metrics, aleatoric uncertainty leads to a more consistent estimator, as indicated by the normalized ANEES values. The state estimator's covariance better captures the estimator's error. Partial aleatoric uncertainty-based outlier rejection (\texttt{U} + \texttt{AORP}) outperforms all other methods across every metric. For example, visualized in \cref{fig:direct_measurement_mug}, the symmetry of the mug causes erroneous rotation predictions, thus motivating partial measurement rejection. Assuming a single object scenario, rejection of the full object-relative measurement leads to prolonged segments of only IMU propagation, leading to dead reckoning and ultimately divergence. However, by only rejecting the rotation measurement, we can still perform EKF updates and include more information in the estimation process. Comparing \texttt{AORP} to \texttt{AOR} highlights that partial measurement rejection prevents the filter from diverging. Disregarding trajectory 3, $\chi^2$\texttt{P} leads to improved performance over $\chi^2$, independent of the underlying measurement noise covariance method. This underlines the need for our newly proposed filter formulation using direct measurements to treat the position and rotation measurements individually and to perform partial measurement outlier rejection. The results show the importance of outlier rejection for AI-based object-relative pose measurements for improved performance and reduced divergence. In contrast to $\chi^2$-based outlier rejection, the \texttt{AOR} approach is likelier to fail due to its thresholding, as an unsuitable choice leads to increased measurement rejections. 

\section{Conclusion}
\label{sec:conclusion}

In this paper, we presented a novel formulation for EKF-based object-relative state estimation and highlighted the benefits of DL-based, dynamic aleatoric uncertainty for state estimation. By deriving the update equations for the direct 6-DoF pose measurement, we can consider the position and rotation measurement independently, reducing the deteriorating behavior of erroneous rotation measurements and enabling partial measurement rejection. Compared to the inverse filter formulation, the Monte Carlo simulations showed the improved performance of our approach, while indicating that both filter formulations perform equivalently well in the absence of noise. Furthermore, we cemented using aleatoric uncertainty as a dynamic measurement noise covariance for DL-based 6-DoF object pose measurements. Besides improving the state estimator's performance and consistency, it removes the need for the time-intensive engineering of a fixed measurement noise covariance that can not capture the situational error characteristics of the AI-based object pose predictor. Combining our novel filter formulation with aleatoric uncertainty-based outlier rejection for partial measurements (\texttt{AORP}) leads to drastic improvements. Future work will investigate the possibility of rejecting individual components of the 6-DoF pose measurement and automatic methods for determining suitable \texttt{AOR}(\texttt{P}) thresholds.
\looseness=-1

%\addtolength{\textheight}{-12cm}   % This command serves to balance the column lengths
                                  % on the last page of the document manually. It shortens
                                  % the textheight of the last page by a suitable amount.
                                  % This command does not take effect until the next page
                                  % so it should come on the page before the last. Make
                                  % sure that you do not shorten the textheight too much.

%%%%%%%%%%%%%%%%%%%%%%%%%%%%%%%%%%%%%%%%%%%%%%%%%%%%%%%%%%%%%%%%%%%%%%%%%%%%%%%%

%%%%%%%%%%%%%%%%%%%%%%%%%%%%%%%%%%%%%%%%%%%%%%%%%%%%%%%%%%%%%%%%%%%%%%%%%%%%%%%%

%%%%%%%%%%%%%%%%%%%%%%%%%%%%%%%%%%%%%%%%%%%%%%%%%%%%%%%%%%%%%%%%%%%%%%%%%%%%%%%%

%\section*{ACKNOWLEDGMENT}

%%%%%%%%%%%%%%%%%%%%%%%%%%%%%%%%%%%%%%%%%%%%%%%%%%%%%%%%%%%%%%%%%%%%%%%%%%%%%%%%
\appendix 

Following the definitions and notation style from \cite{sola2017quaternion}, we define $\mathbf{Q}_\theta,\mathbf{R}_\phi,\mathbf{S}_\eta \in SO(3)$ and $\mathbf{v} \in \mathbb{R}^3$. The following equality holds:
\begin{equation}
    \mathbf{R}_\phi Exp(\delta\theta)  \mathbf{R}_\phi^T = Exp(\mathbf{R}_\phi\delta\theta) \quad .
\end{equation}

The derivatives can then be derived as:
\begin{align}
    &\frac{\delta \mathbf{Q}\mathbf{R}^T\mathbf{S}}{\delta \mathbf{R}} = \frac{\delta \mathbf{Q}_\theta \mathbf{R}_\phi^T \mathbf{S}_\eta}{\delta \phi} \\
    &= \lim_{\delta\phi\to 0} \frac{1}{\delta\phi} ((\mathbf{Q}_\theta (\mathbf{R}_\phi Exp(\delta\phi))^T \mathbf{S}_\eta) \ominus (\mathbf{Q}_\theta \mathbf{R}_\phi^T \mathbf{S}_\eta))\\
    &= \lim_{\delta\phi\to 0} \frac{1}{\delta\phi} Log [ (\mathbf{Q}_\theta \mathbf{R}_\phi^T \mathbf{S}_\eta)^T \mathbf{Q}_\theta (\mathbf{R}_\phi Exp(\delta\phi))^T \mathbf{S}_\eta]\\
    &= \lim_{\delta\phi\to 0} \frac{1}{\delta\phi} Log [ \mathbf{S}_\eta^T \mathbf{R}_\phi \mathbf{Q}_\theta^T \mathbf{Q}_\theta  Exp(\delta\phi)^T \mathbf{R}_\phi^T \mathbf{S}_\eta] \\
    &= \lim_{\delta\phi\to 0} \frac{1}{\delta\phi} Log [ \mathbf{S}_\eta^T \mathbf{R}_\phi Exp(-\delta\phi) \mathbf{R}_\phi^T \mathbf{S}_\eta] \\
    &= \lim_{\delta\phi\to 0} \frac{1}{\delta\phi} Log [  Exp(- \mathbf{S}_\eta^T \mathbf{R}_\phi \delta\phi)] \\
    &= \lim_{\delta\phi\to 0} \frac{1}{\delta\phi} - \mathbf{S}_\eta^T \mathbf{R}_\phi \delta\phi = - \mathbf{S}_\eta^T \mathbf{R}_\phi
\end{align}
\begin{align}
    &\frac{\partial \mathbf{QR}^T\mathbf{v}}{\partial \mathbf{R}} =  \lim_{\delta\phi\to 0} \frac{\mathbf{Q}\mathbf{R}\{\phi + \delta \phi\}^T \mathbf{v} - \mathbf{Q}\mathbf{R}^T\mathbf{v}}{\delta\phi} \\
    & = \lim_{\delta\phi\to 0} \frac{1}{\delta\phi} (\mathbf{Q}(\mathbf{R}Exp(\mathbf{J}_r(\phi)\delta\phi)^T - \mathbf{Q}\mathbf{R}^T)\mathbf{v} \\
    & = \lim_{\delta\phi\to 0} \frac{1}{\delta\phi} (\mathbf{Q}Exp(-\mathbf{J}_r(\phi)\delta\phi)\mathbf{R}^T - \mathbf{Q}\mathbf{R}^T)\mathbf{v} \\
    & = \lim_{\delta\phi\to 0} \frac{1}{\delta\phi} \mathbf{Q} (Exp(-\mathbf{J}_r(\phi)\delta\phi) -\mathbf{I}_3)\mathbf{R}^T\mathbf{v} \\
    & \approx \lim_{\delta\phi\to 0} \frac{1}{\delta\phi} \mathbf{Q} (\mathbf{I}_3 - [\mathbf{J}_r(\phi)\delta\phi]_\mathsf{x} -\mathbf{I}_3)\mathbf{R}^T\mathbf{v} \\
    & = \lim_{\delta\phi\to 0} \frac{1}{\delta\phi} - \mathbf{Q} [\mathbf{J}_r(\phi)\delta\phi]_\mathsf{x}\mathbf{R}^T\mathbf{v} \\
    & = \lim_{\delta\phi\to 0} \frac{1}{\delta\phi} \mathbf{Q} [\mathbf{R}^T\mathbf{v}]_\mathsf{x} \mathbf{J}_r(\phi)\delta\phi \\
    & = \mathbf{Q} [\mathbf{R}^T\mathbf{v}]_\mathsf{x} \mathbf{J}_r(\phi)
\end{align}

\bibliographystyle{IEEEtran.bst}
\bibliography{root.bib}
\end{document}